%% file: eval_avx512.tex
\documentclass[conference]{IEEEtran}
\IEEEoverridecommandlockouts

\usepackage{cite}
\usepackage{amsmath,amssymb,amsfonts}
\usepackage{algorithmic}
\usepackage{graphicx}
\usepackage{textcomp}
\usepackage{xcolor}
\def\BibTeX{{\rm B\kern-.05em{\sc i\kern-.025em b}\kern-.08em
    T\kern-.1667em\lower.7ex\hbox{E}\kern-.125emX}}

\usepackage{hyperref}

\newlength{\halfwidth}
\newlength{\temp}

\begin{document}

\title{Improving a Parallel C++ Intel AVX-512 SIMD\\
Linear Genetic Programming Interpreter}

\author{%
\IEEEauthorblockN{\href{http://www.cs.ucl.ac.uk/staff/W.Langdon}
{William B. Langdon}}
\IEEEauthorblockA{\textit{Department of Computer Science,} \\
\textit{University College London,}\\
Gower Street, London, UK \\
w.langdon@cs.ucl.ac.uk}
}

\maketitle
\pagestyle{plain}

\begin{abstract}
We extend recent 256 SSE vector work to 512 AVX giving a four fold speedup.
We use
\href{https://github.com/bloa/magpie}
{\mbox{MAGPIE}}
(Machine Automated General Performance Improvement via Evolution of software)
to speedup a C++ linear genetic programming interpreter. 
Local search is provided with three alternative
hand optimised codes,
revision history
and the Intel 512~bit AVX512VL documentation as C++ XML\@.
Magpie is applied to the new Single Instruction Multiple Data (SIMD) parallel interpreter
for Peter Nordin's linear genetic programming GPengine.
Linux mprotect sandboxes whilst performance is given by perf instruction count.
In both cases,
in a matter of hours local search reliably sped up 114 or 310 lines of
manually written parallel SIMD code
for the Intel Advanced Vector Extensions (AVX)
by 2\%.
\end{abstract}

\begin{IEEEkeywords}
linear GP, LGP,
genetic improvement, GI,
SBSE,
computer program tuning,
MMX,
transplantation,
testing interpreters,
test output distribution,
non-functional GI,
LUT,
entropy,
dogfooding.
\end{IEEEkeywords}

\section{Introduction}

\noindent
Until about 2005,
with the introduction of 3~gigahertz Intel Pentium~4,
CPU processor speeds had more-or-less
kept pace with Moore's Law~\cite{Moo65},
with clock rates doubling approximately every two years.
If this remarkable achievement of the second millennium
had continued for the last twenty years
we would be running terahertz laptops.
However Moore's Law was originally stated in terms of the number
of components
and to a large extent the exponential increase in the number of transistors
per integrated circuit chip has continued.
Today the Moore's Law bounty is being put to use in parallel computing.
Today typically high performance computers (HPC) are constructed not
from a single rapid computer but from thousands
of networked mundane everyday computers each of which contains a number
of parallel computing elements.
(Often HPC performance is boosted by a small number (1--8)
of parallel GPU boards per computer.)
Concentrating upon the computers,
their CPU chip typically contains a few dozen parallel core
which share memory caches and
implement the same full computing instruction set.
Often they are programmed as if they were complete computers.
The item of interest here,
is both Arm and Intel's instruction sets
have vector operations which process multiple data items in parallel.
For example,
the Intel 512~bit AVX
instruction allows 64 eight bit numbers to be processed simultaneously.
(The ARM instruction set allows longer vectors.)
Since the i7-9800X CPU 3.80GHz supports AVX and we
use GPengine with 8~bit byte data~\cite{Langdon:2026:raLGP},
during manual development
we concentrated upon the 512~bit AVX intrinsics available to C++.

Although it has long been obvious,
baring a room temperature quantum computing break through,
that the future of computing hardware is parallel,
the software tools to support general programming on parallel hardware
are lacking and fully automatic parallel programming has
not been obtained.
Instead manual parallel programming remains hard
and calls for highly skilled specialist programmers.

Rather than relying purely on manual effort,
Conor Ryan~\cite{ryan:1995:paragen,ryan:1999:aigp3}
suggested the use of search based tools in the form of genetic programming
\cite{koza:book,poli08:fieldguide}
to aid the generation of parallel code.
More recently Genetic Improvement~%
\mbox{\cite{Langdon:2012:mendel,Langdon:2013:ieeeTEC,%
Langdon:2016:SSBSE,
Petke:gisurvey,%
blot:2021:tevc,%
bian:2021:apr_icse,%
Guizzo:2024:TOSEM,%
Williams:2024:ICSE-SEIP%
}}
has been used to speedup production parallel code
(e.g.~SSE~\cite{Langdon:2013:ieeeTEC} and
CUDA~\cite{langdon:2014:EuroGP,Langdon:2016:GPEM})
and even to improve evolutionary computing itself
(EC to improve EC!).
For example, using GISMO to speed up
the traditional tree genetic programming systems
Beagle puppy~\cite{Lopez-Lopez:2019:SC}
and GPquick~\cite{langdon:2020:cec}.
Here we apply Magpie to the performance critical component
of linear genetic programming,
towit GPengine's interpreter.
For our Mackey-Glass experiments~\cite{Langdon:2026:raLGP},
GPengine needs to support eight bit,
addition, subtraction, multiplication and protected division.

Genetic programming interpreters
need high performance~\cite{langdon:2022:trillion}
but also require operations to be ``protected''~\cite{koza:book}.
In particular they need to protect division by zero
and so protected division by zero is often defined to give a result:~0,
rather than, for example, throwing an exception%
\footnote{The analytic quotient
AQ($x$,$y$) $= \frac{x}{\sqrt{1+y^2}}$
provides a more continuous alternative to division
\cite{Ni:2012:ieeeTEC}.}.
This simple way of protecting division causes implementation problems with
parallel vector instructions SIMD~\cite{tufts93,pollack:1996:aigp2,langdon:2008:eurogp}
which require all data to be treated in the same way.
In the manually written SSE interpreter~\cite{langdon:2026:GI}
this was eventually resolved by
replacing actual division by looking up the answers in
a 256~by~256 (8~bit by 8~bits, 65\,536) table of precomputed results.
If the look up table (LUT) consists of 32 bit wide entries,
SSE and AVX512 ``gather\_epi32''
instructions can be used to 
lookup the result of 8 or 16 protected divisions simultaneously.
Notice neither SSE nor AVX support integer division.
However protected division is possible via floating point division
and integer truncation.
For example,
Intel's AVX~512 instruction set
includes data dependent vector operations,
e.g.\ \_mm512\_maskz\_div\_ps and \_mm512\_mask\_blend\_ps.
The first can perform division, using a suitable mask
to avoid divisions by zero and then using
the blend instruction to supply any missing results
\cite{langdon:2022:trillion}.
Since the look up table involved complicated indexing operations,
it is was far from clear that the manual code
was optimal and
so it was abandoned in production~\cite{Langdon:2026:raLGP}.
Therefore in Section~\ref{sec:gi_landscape} we turn to Genetic Improvement.

The next two sections describe the history of
our target software: GPengine
and then our ``out of the box'' use of the newest version of Magpie,
particularly setting up the C++ sources it is to optimise as XML files,
test cases for a simple program interpreter
and sandbox hardening the fitness function.
Section~\ref{sec:results} describes
the general code improvements found
and Magpie's performance,
which is further discussed in Section~\ref{sec:discuss}.
In Section~\ref{sec:conclude}
we conclude that Magpie can find correct parallel SIMD speedups
which exploit the available AVX instruction set
and consider alternative future approaches,
including discussing using Magpie for transplantation.

\section{Genetic Improvement of GPengine}
\label{sec:gi_landscape}

\noindent
GPengine was provided by Peter Nordin,
who wrote the super fast commercial
linear genetic programming system Discipulus
\cite{kinnear:nordin,francone:manual,foster:2001:discipulus}).
we had earlier used GPengine 
\cite{langdon:2001:elvis,langdon:2005:CS}
but it had been little used recently.
Nonetheless it is a simple clean C++ implementation
of linear genetic programming and seemed ripe for conversion
to modern parallel computing.

\section{Magpie}
\label{sec:magpie}

\noindent
Magpie 
(Machine Automated General Performance Improvement via Evolution of software)
\cite{blot:2022:corr_1} 
is a development of PyGGI~\cite{An2017aa}
but is language independent.
It 
was first released by Aymeric Blot in 2022%
\footnote{We use Magpie downloaded 5 Nov 2025 
\url{https://github.com/bloa/magpie}.}.

Initial results using just a double precision implementation of
protected division
proved encouraging~\cite{langdon:2026:GI},
with Magpie finding (in context) a wholly correct
replacement of \verb'== 0' by \verb'=> 0'.
Linux perf showed this unexpected but simple substitution saved
one instruction.
In retrospect we can see that the IEEE~754
double precision float occupies two (32~bit) words
and therefor requires two instructions to test for zero.
Whereas the IEEE~754 standard defines a sign bit
(which separates numbers $\ge 0$ from negative numbers
and only a single instruction is needed to test the sign bit.
Surprisingly the GNU C++ compiler shows a similar preference
for \verb'=>' even for 32~bit {\tt int} data
(see Section~\ref{sec:better_code}).

In \cite{langdon:2026:GI} we describe how encouraged by this,
we moved onto Magpie experiments which evolve the whole
of the manually written GPengine SSE interpreter.
When hardware supporting the longer and more 
complete AVX512 Intel intrinsics became available,
we extended our experiments.
AVX supports 64 parallel 8-bit addition and subtractions,
32 16-bit multiplies and
16 32-bit table look ups.
(AVX also support 16-bit and 32-bit
additions, subtractions and multiplications).
It was not clear if we should use 8-bit, 16-bit or 32-bit registers
to simulate the needed 8-bit GP operations.
In the first experiment
we used conditional compilation to support all three,
with Magpie during evolution choosing between them.
Notice the 16-bit and 32-bit versions need to
convert full width calculations to unsigned 8-bit (0--255) results.
Since this requires executing further instructions,
as well as the base additions etc.,
it was not clear which choice would be optimal.
The first experiment (310 lines of code)
confirmed
that 8-bit option was the fastest.
The last experiment removed the 16-bit and 32-bit manual code
(leaving just 114 LOC),
to see if Magpie could find further optimisations.

\subsection{Setting up Magpie: Defaults and The Scenario File}
\label{sec:defaults}

\noindent
Magpie's local search with
100\,000 steps was used on four C++/XML source files
(see next section).
and one parameter file.
The parameter allows Magpie to choose the register width
(8, 16 or 32 bits, default 8 bits) when each mutant is compiled.
The XML edits were:
{\tt
    Srcml\allowbreak{}Arithmetic\allowbreak{}Operator\allowbreak{}Setting,
    Srcml\allowbreak{}Comparison\allowbreak{}Operator\allowbreak{}Setting,
    Srcml\allowbreak{}Numeric\allowbreak{}Setting,
    Srcml\allowbreak{}RelativeNumeric\allowbreak{}Setting,
    Srcml\allowbreak{}Stmt\allowbreak{}Deletion,
    Srcml\allowbreak{}Stmt\allowbreak{}Insertion,
    Srcml\allowbreak{}Stmt\allowbreak{}Replacement,
    XmlNode\allowbreak{}Deletion<stmt>,
    XmlNode\allowbreak{}Insertion<stmt,block>,
    XmlNode\allowbreak{}Replace\allowbreak{}ment<stmt>.
}
Each of these edits generates a new valid XML file,
which Magpie automatically converts into a new C++ source file.
By acting via XML, Magpie ensures,
in most cases, its mutations lead to syntactically correct C++
(see Section~\ref{sec:gcc}).
Otherwise Magpie defaults (e.g.~time outs) were used.

\subsection{XML: Documentation, Revision Histories and Manual Code}

\noindent
Four XML files:
IntrinsicsGuide.cpp.xml, diffs.cpp.xml, eval\_diffs.cpp.xml
and eval.cpp.xml, 
were automatically generated by srcml version~1.0.0.
Using the new Magpie {\tt ingredient\_files} option,
the first three are read only and are used as a feedstock for Magpie,
whilst the last contains the interpreter
(i.e.\ the target SUT itself)
and Magpie modifies its XML file
before generating mutated C++ code and attempting
to compile and run it on four test programs.

\subsubsection{Intel IntrinsicsGuide}

\noindent
IntrinsicsGuide.txt%
\footnote{%
\url{https://software.intel.com/sites/landingpage/IntrinsicsGuide/\#}
26 Jan 2017}
documents Intel's C++ runtime library
to support its (assembler based) AVX instruction set.
It is plain text and was automatically converted into C++ code
For example, the sixteen $\times$ 16-bit 
pabsw instruction is documented as
\verb'__m128i _mm_abs_epi16 (__m128i a)'
which is automatically converted to the C++ code
\verb'_mm_abs_epi16 (a);'
IntrinsicsGuide.cpp contains 
4893 functions
plus \_MM\_SHUFFLE.

\vspace{1ex}
\_MM\_SHUFFLE was included via the code
\verb'unsigned char a = _MM_SHUFFLE (z, y, x, w);'

\vspace{1ex}
For all four C++ source files,
comments, assert statements,
empty lines and trailing spaces were removed,
and tabs converted to spaces,
and then srcml was used to analyse the syntax of each,
generate the corresponding abstract syntax tree (AST)
and finally convert it to XML\@.

\subsubsection{GPengine Interpret64}

The
SSE~256 C++ code for Interpret16 and its supporting functions 
from~\cite{langdon:2026:GI}
were rewritten,
keeping the previous SSE structure,
to use AVX512 to give
8, 16 and 32 bit versions (separated by conditional compilation)
of the new function Interpret64.
(Interpret16 evaluated 16 test cases in parallel,
whereas Interpret64 evaluates 64 test cases in parallel.)
This took one week.
In several cases C++'s ability to overload functions
with different arguments was used.
For example, three versions of {\tt \_\_m512i InstrArg16()},
which retrieves 16 data values for the current instruction's OP code
and packs them into a 512 bit vector,
were written for each {\tt registers} type:
{\tt uint8\_t},
{\tt uint16\_t} and
{\tt uint32\_t}.
They each call {\tt InstrArg()},
which was unchanged from the SSE code.
The C++ compiler is happy to compile all three and,
according to the register width Magpie select at compile time,
only generate calls for the one actually needed by the mutant.
Similarly the existing SSE structure was used for multiple version of 
{\tt InstrArg16} and {\tt InstrArg32}
and the {\tt InstrReg*} family of functions.
{\tt InstrArg*} and {\tt InstrReg*}
both return data for the current instructions input registers.
{\tt InstrArg*} are marginally more complicated
as they deal with both data in the given register or the given constant,
whilst {\tt InstrReg*} only read from the register.
They both (if need be) convert from the register format
to the appropriate data packing required for
{\tt \_\_m512i} vectors.

For simplicity, for speed of development and
in case Magpie and the optimising compiler could exploit it,
no attempt was made to manually take advantage of the commonality
between the support functions.
Similarly, in the first experiment,
unused SSE 256 code was left in place
whereas in the second experiment everything
not needed for the 8-bit version was removed.
Thus in the first experiment the volume of code Magpie is free to optimise
(310 lines of code)
is much bigger than in the second (114 LOC).

\subsubsection{GPengine Revision History}

\noindent
Although it is quite common for software engineering experiments
to consider commits and revision metadata,
this seems rare in genetic improvement~%
\mbox{\cite{langdon:2019:EuroGP,%
2007.06986}}. 
During manual development of the SSE version of GPengine's
interpreter 24 
snapshots had been saved into RCS~\cite{Tichy:1982:ICSE}
over 10 days
(note GPengine has more than 25 years of development history).
Each change was automatically extracted
and split into individual changes
(89 in total,
\mbox{1--54} lines each, median~1).
Comments and assert statements were again removed
and empty files were removed leaving
69 C++ files
of between~1 and 51 lines (median~2).
These were concatenated in order,
giving a single C++ file composed of the individual fragments.
srcml was again used, creating a single file, diffs.cpp.xml,
holding all the code changes as XML\@.

\subsubsection{Interpret64 Revision History}

\noindent
As in the previous section,
we gave Magpie access to the manual development history
of the conversion of the interpreter from SSE~256
to AVX512.
During AVX512 development
13 snapshots of eval.cpp were taken,
giving 70 individual changes
(1--16 lines each, median~1).
Again comments, asserts and empty files were removed leaving
54 C++ files
(1--16 lines, median~1).
These were concatenated in order and
processed by scrml,
to give a second history file:
eval\_diffs.cpp.xml

\subsection{Magpie Fitness Function}
\label{sec:fitness}

\subsubsection{Two Compilations}

\noindent
During earlier work \cite{langdon:2026:GI},
we had notice although Magpie mutant's fitness was stable
recompilation with a different g++ command line
could give a program which failed at run time.
In some cases this was traced to Magpie mutating the code
to give an undefined result,
which with command line option -O3,
the compiler optimised away,
whereas without -O3 the resulting program gave a segmentation error
(SegFault).
In the hope of avoiding unstable mutations,
we compile the mutant twice,
the first time without -O3.
If no compilation or runtime errors are reported,
it is re-compiled with -O3 and run again by the
fitness test harness.
(Notice some earlier BNF grammar approaches
ensured all mutants compiled cleanly,
e.g.~\cite{langdon:2009:TAICPART,langdon:2010:cigpu}.)

As Tables~\ref{tab:count_gcc_errors32}
and~\ref{tab:count_gcc_errors8}
will show,
many XML based mutations fail to compile
due to variables being out of scope.
Therefore we attempted to use GCC compilation error messages
which suggested alternatives identifiers
``did you mean xyz?''
to automatically replace the erroneous variable name
with the one suggested by the g++ compiler.
Only in one case in many thousands of mutants
did the suggest change result in the new code compiling.
Even then, this compiled mutation was not useful
and this attempt at automatic fixup was abandoned.
Notice however that Marginean~\cite{Barr:2015:ISSTA}
showed that genetic programming
can be used to successfully fixup variable names
when code is transplanted from donor software to
a new host program.

As mentioned above,
in the first experiment Magpie chooses the size of the register data
from the three supported options (8, 16 or 32~bits) [default~8].
The mutant is compiled up to two times with
this size set via conditional compilation.

\subsubsection{Checking for Equivalent Mutations}
\label{sec:equiv_mutants}

Originally \cite{langdon:2026:GI},
the first part of the fitness function
after warmup,
\label{p.warmup}
XML changes which make no difference to the source code,
e.g.\ replace a value with an identical value,
were rejected.
This was easy to do as we also had to check that Magpie
did not change the read only input files
(IntrinsicsGuide.cpp diffs.cpp).
Since the new version of Magpie supports {\tt ingredient\_files},
we no longer had to do this.
Instead we now check that the compiler
output (.o~file) has changed.
This can be, as it is here, a
highly effective way to detect equivalent mutations
(non code changes)
\cite{Papadakis:2015:ICSE,%
langdon:2020:cec,langdon:2023:EuroGP}.
Unfortunately as Magpie's list of mutations grows,
checking the resulting object file
against the unmutated code,
is unable to reject the latest addition
even if adds nothing.
It is feasible to maintain a complete tabu list of
equivalent mutations 
\cite{langdon:2018:EuroGP},
but only a tabu of one edit was enforced here.

\begin{table*}
\caption{Test input data for the four test programs
(64 pairs of $x/y$ inputs per program)
\label{tab:eval2latex}
}
\vspace{-4ex}
\begin{center}
\input{graph512/eval2latex.tex}
\end{center}
\end{table*}

The evolved code is compiled separately from the test harness
and then 
they are linked together as a linux exe file
which is run by Magpie.
We use the GNU C++ compiler (11.5.0),
with optimisation \verb'-O3',
\verb'-fmax-errors=1'
and for our version of AVX (\verb'-march=skylake-avx512').
Changes which failed to compile are rejected
Tables~\ref{tab:count_xml32},
\ref{tab:count_xml8},
\ref{tab:count_gcc_errors32}
and~\ref{tab:count_gcc_errors8}).

\subsubsection{Creating Test Cases, Edge Cases and Uniform Output Distribution}
\label{sec:testcases}

\noindent
We randomly create four linear GPengine programs
each composed of four instructions,
Figure~\ref{fig:eval_progs}.
Each is given 64 test cases to be processed in parallel,
i.e.\ 256 test cases in total.
As we anticipated the most problems with protected division,
we insist that all four programs start with division.
The three remaining instructions in each program are selected at random.
However in the third program we insist,
that in total across all four programs,
there is at least one of each of the four possible arithmetic operations
($+ - \times$ and protected division).

\begin{figure}
\begin{center}
\input{graph512/eval_progs.tex}
\end{center}
\vspace{-2ex}
\caption{Four GPengine test programs each with four instructions
\label{fig:eval_progs}
}
\end{figure}

GPengine's instructions comprise an opcode \mbox{($+ - \times$ or $/$)},
an output register
and two inputs.
One input is always a register
and the second is either a constant or a register.
There are 8 registers.
For each of the four programs we choose uniformly at random
two input registers and an output register.
The first instruction uses as input the two input registers
and uniformly at random chooses its own output register.
(All three registers are randomly chosen
and so are free to overlap.)

The second and third instructions similarly randomly choose their
output registers.
However they are forced to choose their input registers from registers with known values,
i.e.~those previously written to or the two program input registers.
However the opcode's second input is chosen like GPengine does,
i.e.\ a fraction (20\%) are one of these registers with a known value
and the rest are constants (uniformly chosen from the range 0 to~127).
The last instruction is the same,
except its output is forced to be the program's chosen output register.

\begin{figure}
\begin{center}
\includegraphics{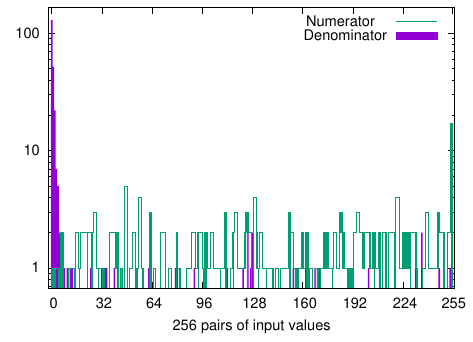}
\vspace*{-2ex}
\caption{Distribution of 256 pairs of fitness input values.
Notice concentration at edge cases 0, 1 and 255
(log vertical scale).
\label{fig:inputs}
}
\end{center}
\end{figure}

\begin{figure}
\begin{center}
\includegraphics{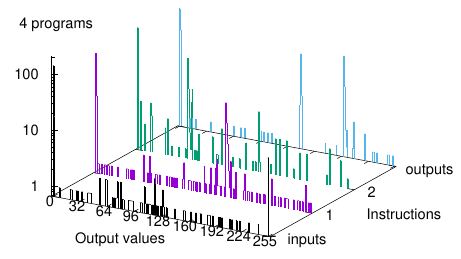}
\vspace*{-2ex}
\caption{Sum of 
distribution of 256 output values across four test programs
during fitness testing.
Starting at the input
(instruction 0,
protected division, see Figure~\protect\ref{fig:inputs}),
instructions 1 and 2 and values output by the 4 programs.
(log vertical scale).
\label{fig:out4}
}
\end{center}
\end{figure}

\begin{figure}
\begin{center}
\includegraphics{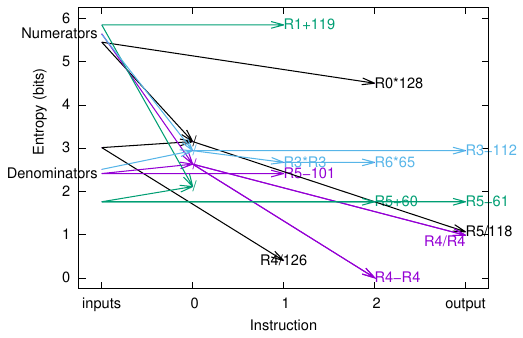}
\vspace*{-2ex}
\caption{Entropy of
distributions of 64 values for each of the four test programs
(1~black, 2~purple, 3~green and 4~light blue).
Starting at the inputs
(protected division, Figure~\protect\ref{fig:inputs}),
instructions 1 and 2 and values output by the 4 programs
(Figure~\protect\ref{fig:out4}).
Note due to {\tt int} wrap around
\mbox{\tt + -} tend to loose little information,
compared to multiply and protected division \mbox{\tt * /}
\label{fig:entropy}
}
\end{center}
\end{figure}

Care was taken with the first instruction's (protected division)
data values.
The results of that instruction may be
propagated to the remaining three instructions.
To force execution of all paths;
uniformly at random half the data forces division by zero,
Table~\ref{tab:eval2latex}. 
Half the remaining 50\% are chosen to 
force edge cases.
One in eight pairs of input values are chosen to give output of 0,
of 1, of 255 and a random value between 1 and 255.
The remaining four in eight (i.e.\ 25\% of all test cases)
are chosen to give an output uniformly chosen between
2 and 127,
cf.~\cite{Menendez:2022:TSE},
Figure~\ref{fig:inputs}.
Figure~\ref{fig:out4} plots the distribution of the outputs
produced by the program instructions,
whilst Figure~\ref{fig:entropy}
shows the entropy of the distributions split into
the values produced by the four test programs.

\subsubsection{Sandboxing mprotect 4KB Pages}
\label{sec:sandbox}

\noindent
``Sandboxes'' are techniques to limit the damage that
running random code might cause.
By default Magpie provides some limited protection against evolved
code running amok.
These are chiefly:
timing out evolved code that is stuck in indefinite loops
and by running it in separate processes and thus
taking advantage of the operating system's protection.
(We use Linux Rocky~9.6.)
Note eval does not contain file, I/O statements
or system calls,
which also limits the scope for damage.

Early eval\_avx experiments
showed  mutant code writing to array index -1,
i.e.\ outside the legitimate range of the array,
which C++ does not forbid
(it is undefined).
Early grammar based GI approaches enforced array bound checks (%
e.g.~\cite{langdon:2010:cigpu}
or used hardware support, e.g.~\cite{Langdon:2014:GECCO}) 
common in modern programming languages
but absent from C++
\cite{langdon:2020:cec}.
Instead we use Linux mprotect to provide guards
around data which is given to the evolved code.
It appears that this and~\cite{langdon:2026:GI}
are the first time the mprotect
mechanism has been used with Genetic Improvement.

\begin{figure}
\begin{center}
\centerline{
\includegraphics[scale=0.25]{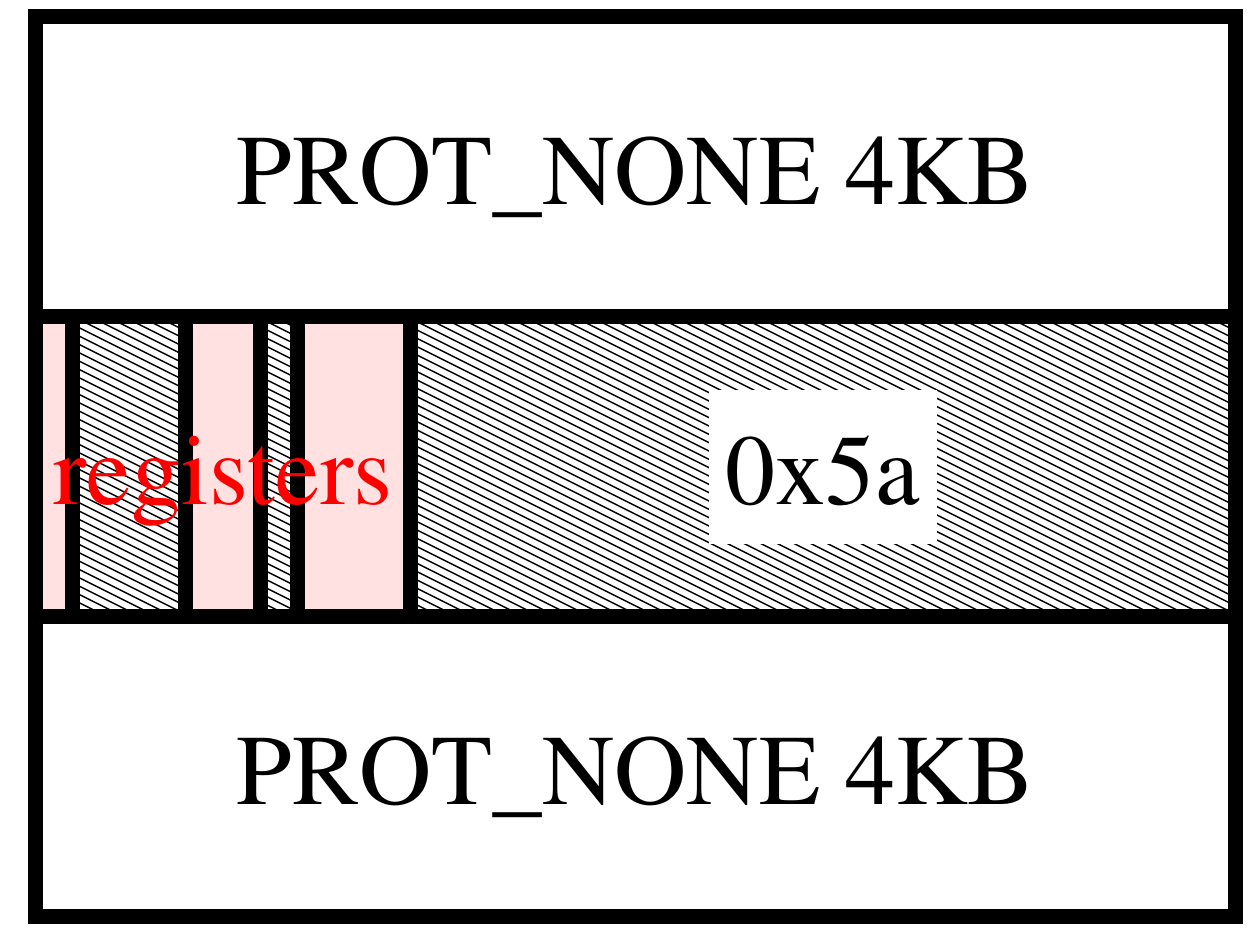}}
\caption{The read-write I/O registers are surrounded by 4K byte buffers
where either read or write access
will cause an illegal access violation SegFault.
Unused bytes are filled with 90 ($5\times 16 +10$, Z,
4 bits set 4 bits clear).
After running the mutant,
the test harness checks the pattern has not been disturbed.
The division look up table and the program are
also given PROT\_READ
as well as being similarly surrounded by 4KB guards.
\label{fig:mprotect}
}
\end{center}
\end{figure}

When the evolved code (Interpret64) is called
by the C++ test harness,
it is past the length of the program,
the program,
the 8 registers (with the inputs preloaded)
and the protected division lookup table.
I.e., three arrays, one array is read/write (the registers)
and two should be read only.
Linux (Rocky 9.6) mprotect works on 4KB memory pages.
Each of our three arrays is forced to start at a 4KB boundary
(Figure~\ref{fig:mprotect}).
Either side of each array the test harness declares empty
4KB arrays, which it uses mprotect to disable any access to.
Thus if the evolved code attempts to access array element~-1
the operating system will issue a segmentation error (SegFault)
and the test harness will be aborted
and Magpie will treat this as a failure to set a fitness
and move on to generate the next code mutation.
Also the pages holding
the program and the lookup table are protected to allow only read access.
To simplify the test harness,
and avoid a SegFault on {\tt main() return},
rather than undoing all the mprotect calls,
the test harness simply uses the Linux process exit routine directly.

\label{p.sandbox_padding}
The I/O register array
does not fill a complete 4KB window.
Therefore it is padded up to 4KB
and the unused memory is loaded with a
non obvious data pattern.
(The same pattern is loaded into all the registers
except those holding the test program's inputs.)
After each time the evolved code finishes,
the test harness checks that the padding pattern
and registers which the program should not use
have not been changed.
Obviously this cannot check if a mutation
read memory inside the 4KB window it should not have,
but write access is likely to be detected
(Status~3 and~4 in
Tables~\ref{tab:run_time_errors32} and~\ref{tab:run_time_errors8}).
Like SegFaults this is treated as a fatal error,
the test harness stops immediately and
Magpie does not assign the mutant a fitness.
(No special care is taken to pad the area occupied by the four programs
to the next 4KB boundary.
Whereas the look up table fits neatly into 64 4KB pages.)

This protection seems to be good enough.
It is not 100\% fool proof.
For example, it only protects memory address,
not array indexes.
So, for example, multi dimensional arrays such as the test programs
and the lookup table
have multiple indexes:
any small misuse of these is liable to incorrectly access
a different part of the array,
which to the operating system, will appear as a legitimate
address within the array and no SegFault will be issued.
Similarly a large addressing error may step over the 4KB
protection windows, possibly into an unprotected random part of the 
test harness.

When the evolved code terminates,
all 8 registers are checked.
The test harness checks
that those registers which the test program should not have used
still contain the padding pattern.
(Any failure here aborts fitness testing, see above).
It checks the values in all the other registers
(even if they should only contain the results of intermediate calculations).

For each used register (for each of the 64 test cases)
the error is the absolute value of the difference between its value
and the expected value
(ideally there should be no difference, i.e.\ the error should be zero).
These are summed.
The error used is the sum over the four test programs.
If this is not zero a large value is added
which ensures erroneous mutants always have fitness bigger (i.e.\ worse)
than a correct mutants no matter how slow it is.

As recommended by Blot and others
\cite{%
Blot:2020:EuroGP,
blot:hal-03595447,
langdon:2024:ASE,
bouras:2025:ssbse}
we use Linux perf 
to gather statistics on run time,
and use perf's instruction count
(\url{https://github.com/wblangdon/linux_perf_api})
rather than elapsed time
as it is far more stable.

\section{Results}
\label{sec:results}

\noindent
Table~\ref{tab:count}
summarises the outcome of the five Magpie runs
in each of the two experiments.
The next section describes in detail the fastest of
the
(average per run)
65\,053 correct mutants in the first experiment
and the 15\,687 in the second experiment.
In the second experiment,
where Magpie is given the best register size (8~bits)
from the first experiment,
all the runs again find the same three code improvements.

\begin{table}
\caption{Mean out come of 100\,003 mutants across five Magpie runs.
Columns 2--4 first and second (col~5) experiments.
\label{tab:count}
}
\begin{center}
\input{graph512/count}
\end{center}
\end{table}

\subsection{Code Changes and Fitness Improvement}
\label{sec:better_code}

\begin{figure}
\begin{center}
\includegraphics{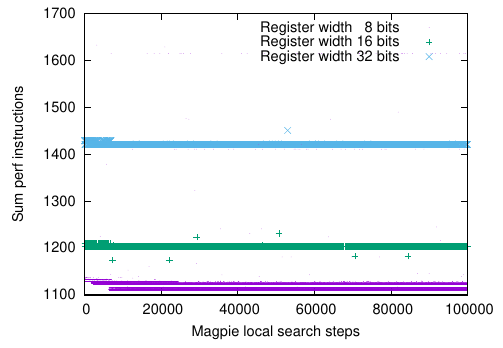}
\vspace*{-2ex}
\caption{Speed of correct Magpie mutants
in five runs with choice of register size.
8~bits (320\,656~ok mutants 1111--1633, median~1111),
16~bits (2\,367~ok mutants 1175--1231, median~1203) and
32~bits (2\,252~ok mutants 1421--1451, median~1421).
\label{fig:fitness_32}
}
\end{center}
\end{figure}

\begin{figure}
\begin{center}
\includegraphics{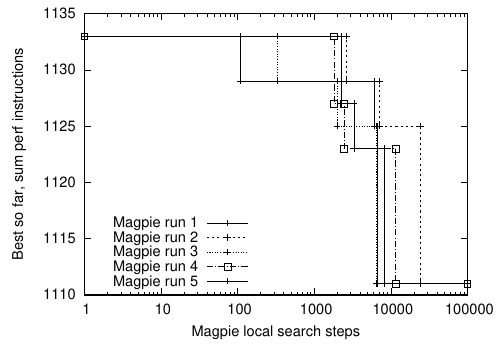}
\vspace*{-2ex}
\caption{Best fitness in 5 Magpie 
runs with choice of register size,
log x-scale
\label{fig:fitness_32a}
}
\end{center}
\end{figure}

\begin{figure}
\begin{center}
\includegraphics{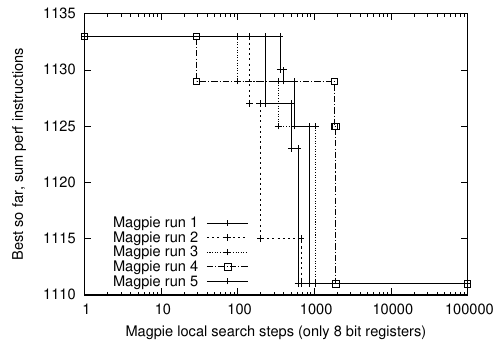}
\vspace*{-2ex}
\caption{Best fitness in five Magpie 2$^{\rm nd}$ runs.
Note log x-scale.
\label{fig:fitness_8}
}
\end{center}
\end{figure}

\begin{figure}
{\tt \flushleft \small
XmlNodeReplacement<number>(('eval.cpp.xml', 'number', 9), 
\mbox{('eval\_diffs.cpp.xml', 'number', 38))} |
\mbox{SrcmlComparisonOperatorSetting(('eval.cpp.xml',} 
\mbox{'operator\_comp', 4), '>=')} |
\mbox{SrcmlNumericSetting(('eval.cpp.xml', 'number', 4), '0')}
}
\vspace*{-3ex}
\caption{
Best solutions.
Magpie XML patch.
It takes 1111 instructions to execute the test cases.
All runs found this speedup.
It comprises three edits (separated by vertical bars~$\vert$~). 
{\tt XmlNodeReplacement<number>}
and
{\tt SrcmlNumericSetting}
both replace numeric constants {\tt 0xaaaaaaaaaaaaaaaa}
with {\tt 0}.
{\tt Srcml\allowbreak{}Comparison\allowbreak{}Operator\allowbreak{}Setting}
replaces {\tt operator\_comp} number~4
which is a {\tt ==}
by a {\tt >=}. 
\label{fig:1111_XML}
}
\vspace{2ex} 
\includegraphics[width=\halfwidth]{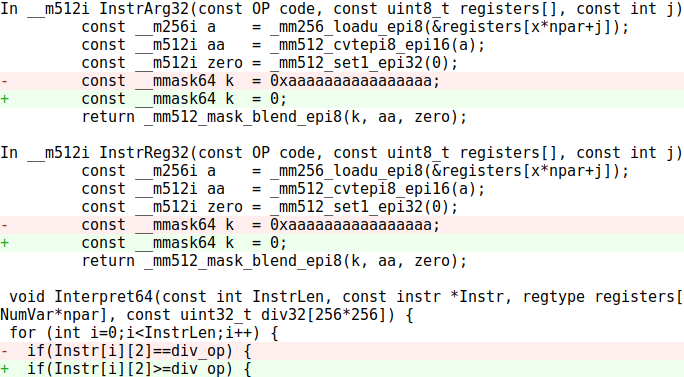}
\vspace*{-3ex}
\caption{
The best solution makes three C++ code changes
(see also Figure~\ref{fig:1111_XML}).
In both the support routines
the mask to ensure
that the 16~bit version of the 8~bit register is kept in $0\cdots 255$
is disabled.
The last change
is at top of the for loop which steps through the program
and replaces {\tt ==} by $\ge$.
\label{fig:1111_patch}
}
\end{figure}

\noindent
Figures~\ref{fig:fitness_32}, \ref{fig:fitness_32a} and~\ref{fig:fitness_8}
shows the evolution of the best fitness,
i.e.~the sum of the number of instructions used
by the four test programs
in five independent runs.
In both experiments all five runs found the same solution,
see
Figures~\ref{fig:1111_XML}
and~\ref{fig:1111_patch}.

The code changes and speed up are described in
Figures~\ref{fig:1111_XML}
and~\ref{fig:1111_patch}.
It resembles the best SSE~256 mutant~\cite{langdon:2026:GI}.
They are ok since:

The first two Magpie edits are both in code which reads
32~byte sized data values from either the first or
the second half the {\tt x}$^{\rm th}$~register
(depending on {\tt j}).
It then sign extends them to 16~bits
(thus filling the 512~bit vector ({\tt aa}).
This is needed as the following operations (not shown)
work on 16~bit components of {\tt \_\_m512i} vectors.
The sign extension was not wanted
and so the following {\tt \_mm512\_mask\_blend\_epi8}
is needed to ensure the top byte of each 16~bit
element is zero.
However both support routines are only used by the
{\tt mul\_op:} multiply code,
which takes two vectors of 16~bit data and
after multiplying them converts their product
back 8~bits
removing the unwanted upper byte.
It appears Magpie and the O3 optimising compiler
spotted that the {\tt \_mm512\_mask\_blend\_epi8}
operation is not needed and by replacing
{\tt k} with zero,
the compiler can remove it and so speed up the code.

For the last Magpie edit:
since div\_op is the largest opcode,
replacing an equality test by a $\ge$ makes no difference to the
{\tt Instr[i][2]}~vs.~{\tt div\_op} comparison.
However it is marginally faster.

The peak speed of the evolved GPengine interpreter
(excluding crossover mutation etc.)
is $4 \times 4 \times 64 \times 3.80{\rm GHz} /1111 $
= 3.5 billion GP operations per second (3.5 Giga GP/s),
i.e.\ 3.9~times faster than the SSE 256 version~\cite{langdon:2026:GI}.

\subsection{Size of Edits; Number of Mutations}

\noindent
In our first five runs,
16\%
of Magpie mutations are not new but instead
it is trying again a mutation and
so does not recalculate its fitness but instead
pulls it from its cache
(top row in Table~\ref{tab:count}).
For the second experiment it is 32\%. 
\label{p.cache}
Excluding those cached,
mutations typically contain between 1 and 34 individual edits
(mode~7)
with on average Magpie concentrating only
29\% of its new trials on
the most popular lengths
(typically 6--8 edits).

For the second experiment
(i.e.\ with the register size fixed at 8~bits
and so no edits to tune it),
Magpie homes in on the three edit best solution faster
and makes more use
(32\%) 
of its fitness cache of known mutations.
Its mutations are shorter
(mode~4)
and more tightly bunched,
e.g.\ 87\%
are one of the popular length (4--6).

\begin{figure}
\includegraphics{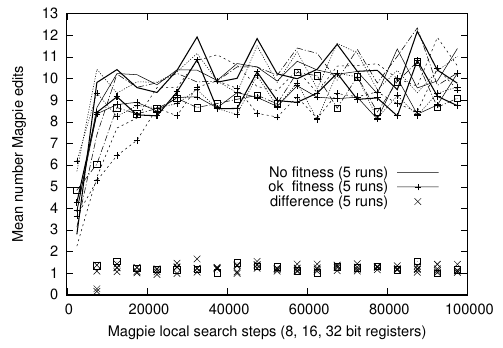}
\vspace*{-2ex}
\caption{Mean number of edits in five Magpie runs
(averages of 5000 steps).
Lower crosses ($\times$) show on average the difference between edits
which failed fitness testing (no tick marks)
and those with fitness (lines with tick marks) is $\approx\!\!1$.
\label{fig:lengths_32}
}
\end{figure}

\begin{figure}
\includegraphics{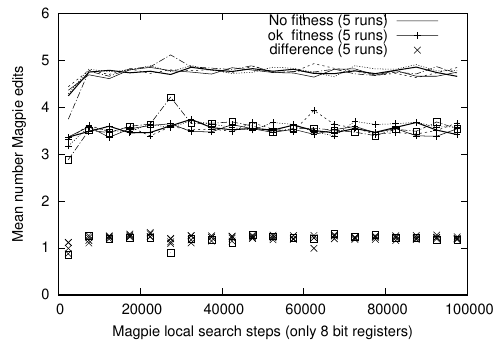}
\vspace*{-2ex}
\caption{Mean number of edits in five Magpie runs
(averages of 5000 steps
as Figure~\protect\ref{fig:lengths_32} but Magpie 
can only use register size 8).
Lower crosses ($\times$) show on average the difference between edits
which failed fitness testing (no tick marks)
and those with fitness (with tics) is $\approx\!\!1$.
\label{fig:lengths_8}
}
\end{figure}

Figures~\ref{fig:lengths_32} and~\ref{fig:lengths_8}
shows the average length of Magpie edits as it searches.
They include Magpie's use of its cache
but the data are split into mutants which failed (top)
and those which ran ok and produced the right answers
(lines with with tick marks).
The local search strategy appears to give rise to
most additional edits
(i.e.~increase length by 1)
failing either compilation or run time checks
(top lines in Figures~\ref{fig:lengths_32} and~\ref{fig:lengths_8}).
The difference between top and bottom lines 
in Figures~\ref{fig:lengths_32} and~\ref{fig:lengths_8}
($\approx\!\!1$)
can be explained by
Magpie's local search strategy, which both randomly adds 
and deletes edits from its active search point.
However,
as our edits appear to be somewhat independent,
random removal is likely to yield an ok mutant
compared to a random additional edit.

\subsection{Success of Mutations}

\noindent
Table~\ref{tab:count_xml32}
shows
only about a quarter of new edits are rejected before or during runtime.
This may be because (e.g.~due to conditional compilation)
many mutation land in code which is not compiled
or not executed.
As these have no fitness impact,
they may also be responsible for increasing
the length of Magpie edits
(previous section).
Perhaps also due to the volume of non-compiled code,
code donated via
the two revision histories and the AVX documentation
is also often successful,
but in the end is never included in the best mutation
(Section~\ref{sec:better_code} above).
Similarly, although Magpie does explore code mutations
for 16 and 32~bit registers,
this never does better than the best 8~bit code
(see Figure~\ref{fig:fitness_32}).

For the second experiment
Table~\ref{tab:count_xml8}
shows a reversed pattern,
with approximately three quarters of new mutations being
unsuccessful.
Note in the second group of five runs
there is no conditional compilation
and so all parts of the code should be compiled
and run.
This may also be why Magpie mutations are much more
focused around the best solution.
Again Magpie does explore code from
the two revision histories and the AVX documentation.
Although these mutations are sometimes successful,
again they are not incorporated into the best code
(see Figure~\ref{fig:fitness_32}).

\begin{table}
\caption{Mean sources of changes across five runs
with Magpie choosing register sizes.
Column~8 gives the total for that row.
\label{tab:count_xml32}
}
\begin{center}
\input{graph512/count_xml32.tex}
\end{center}
\vspace{1ex} 
\caption{Mean percentage sources of changes across five runs
with byte sized registers.
Column~8 gives the total for that row.
\label{tab:count_xml8}
}
\begin{center}
\input{graph512/count_xml8.tex}
\end{center}
\end{table}

\subsection{Compilation Errors}
\label{sec:gcc}

\noindent
As Tables~\ref{tab:count_gcc_errors32}
and~\ref{tab:count_gcc_errors8}
give the compilation errors encountered in five Magpie runs
in the two experiments.
They show most compilation errors
are related to problems with variable names~({\em id}).
Less than 5\% are due to syntax errors.

\begin{table}
\caption{Break down of 69\,170 Compilation errors across five runs
with Magpie tuning the register size
\label{tab:count_gcc_errors32}
}
\setlength{\tabcolsep}{2pt}
\vspace*{-1ex}
\input{graph512/count_gcc_errors32.tex}
\end{table}

\begin{table}
\caption{Break down of 148\,404 Compilation errors across five runs
register size 8~bits.
\label{tab:count_gcc_errors8}
}
\setlength{\tabcolsep}{2pt}
\vspace*{-1ex}
\input{graph512/count_gcc_errors8.tex}
\end{table}

The compilation process
(including checks that the object file is indeed different
Section~\ref{sec:equiv_mutants} above)
never timed out.
Typically it takes less than a second.
(As mentioned in Section~\ref{sec:defaults},
we use the default Magpie timeout,
30~seconds.)

Although discarding faulty mutants due to compilation errors
is often cheap compared to finding faults by running them,
the high fraction rejected by the compiler suggests
our text based conversion to XML
is too simplistic.
Transplantation work by Alexandru Marginean~%
\cite{%
Barr:2015:ISSTA,%
Marginean_10137954_thesis_redacted}
showed that search in the form of genetic programming
can be used to fix up variable name differences between
the donor code (here IntrinsicsGuide, diffs.cpp and eval\_diffs.cpp)
and the host (eval.cpp).
Alternatively, earlier work
\cite{Langdon:2017:GI,langdon:2019:EuroGP,langdon:2020:cec}
automatically extracted type information, e.g.~from the Intel documentation,
or variable scope information
and incorporated it into a grammar
and used the grammar to enforce type constraints.
Additionally the variable scope information can be used to constrain
code movement (genetic) operations
to ensure variables do not go out of scope.

\subsection{Runtime Errors}
\label{sec:runtime_err}

\noindent
Tables~\ref{tab:run_time_errors32} and~\ref{tab:run_time_errors8}
give summaries of errors detected after compilation
during fitness testing
in five Magpie runs in each of the two experiments.
Column~3 gives the sum across the five runs,
whilst Columns~4 and~5 give means.
Column~4 gives the run time ratios
as fractions of all the
mutants which compiled ok and passed
the check for non-equivalent object code,
Section~\ref{sec:equiv_mutants}.
Whereas Column~5 gives the number of each type of error as a fraction
of all the mutants which failed at run time.

Table~\ref{tab:run_time_errors32} shows in the first experiment
many more mutants pass all the tests than in the second
(Table~\ref{tab:run_time_errors8}).
However this appears to be due to there being many more
equivalent mutants,
which pass tests (due to changes being in non-executed code)
than in the second experiment 
(where all code should be executed).
If we consider the fifth column
(ratios of each type of error) the two experiments are similar.

Most run time errors (61\%,64\% status~1, second row of
Tables~\ref{tab:run_time_errors32},\ref{tab:run_time_errors8})
are caused by mutants
returning one or more wrong answers.
26\% (status~139)
of run time errors
are SegFaults,
some of which are 
illegal reads or writes detected by mprotect
(described above in Section~\ref{sec:sandbox}).
SIGFPE,
SIGABRT,
EBADF,
and SIGEMT,
are described in the tables.
The status~4 and~3 errors
indicate illegal writes by mutants
(described in Section~\ref{p.sandbox_padding}).

\begin{table}
\caption{349\,414 run time errors across five runs with Magpie tuning regsize
\label{tab:run_time_errors32}
}
\vspace{-2ex}
\setlength{\temp}{1.7in}
\setlength{\tabcolsep}{3pt}
\input{graph512/run_time_errors32.tex}
\vspace{-2ex}
\end{table}

\begin{table}
\caption{190,522 run time errors across five runs with fixed regsize=8
\label{tab:run_time_errors8}
}
\vspace{-2ex}
\setlength{\temp}{1.7in}
\setlength{\tabcolsep}{3pt}
\input{graph512/run_time_errors8.tex}
\vspace{-2ex}
\end{table}

There are no run time timeouts.
(Again the Magpie default, 30~seconds, was used).

\section{Discussion}
\label{sec:discuss}
\vspace{-2ex}

\subsection{Testing}

\noindent
Even with Magpie's cache,
on average it takes more than 0.5 seconds
to generate,
check,
compile
and test
each mutant.
Most of this is consumed by the optimising compiler.
In contrast running the fitness test harness
typically takes about 5 milliseconds,
although a few erroneous mutants take much longer.

\subsection{Test Suite Effectiveness}

\noindent
Although we test only four short randomly created GPengine programs
(Figure~\ref{fig:eval_progs}),
with $4 \times 64 = 256$ $x,y$ pairs of inputs
(Table~\ref{tab:eval2latex}), 
these are responsible for eliminating 61\%,64\% of erroneous runable mutants
(top of Tables~\ref{tab:run_time_errors32},\ref{tab:run_time_errors8}).
It seems the forced use of all paths through the interpreter
and wide range of {\em output} values
\cite{Menendez:2022:TSE,Guizzo:2024:TOSEM}
has been effective at ensure mutants which pass the test cases
are indeed correct.

\subsection{mprotect}

\noindent
The Linux mprotect system routine gives an efficient way
of eliminating some badly behaving C++ mutants
whilst also providing a degree of runtime sandbox protection.
In principle mprotect could be extended to cover all of the test harness
but this raises of the practical issues of
turning it off again, without causing a SegFault,
when the mutated code returns control to the test harness
and making reasonable assumptions about how the
optimising compiler will layout its use of memory.
Nevertheless the mprotect windows either side of critical
data structures appears to efficiently detect
about three quarters
of array index corruption issues without impacting
perf's measurement of runtime overhead.

\section{Conclusions}
\label{sec:conclude}

\noindent
Although the importance of parallel programming has long been recognised,
in general efficient programming of vector computing remains
practically impossible for the ordinary human programmer.
Nevertheless
in less than a day Magpie consistently found
small compact non-obvious, comprehensible and correct improvements
(Figure~\ref{fig:1111_patch})
to performance critical
parallel vector code,
which had taken a few weeks to write by hand.

The Linux tools mprotect and perf worked well
giving efficient clean and stable performance measures
leading to code changes we can be confident in.

The improvements were found by local search.
They are independent,
in the sense that each part of the Magpie mutants,
consistently gives an edit responsible for each
and those edits consistently give its own explainable
improvement.
It may be broader, less focused forms of search,
perhaps population based,
such as genetic programming,
might be able to find other improvements by
recombining useful components.
Possibly progress might be made by seeding the initial population~%
\cite{langdon:2000:seed}
with error free (but slower) mutants found by
our local search.
However strong diversity measures
(perhaps fitness sharing \cite{goldberg:book,McKay:2000:GECCO}
or structured populations \cite{Fernandez:2003:GPEM,langdon:2008:eurogp})
might be needed to slow convergence
\cite{langdon:book,langdon:GPEM:gpconv}
and so allow time for effective mixing of partial solutions
from different components.

Allowing search to expand to the whole of the
Intel AVX library
did not bring rewards.
We feel this needs more care;
firstly to ensure
that their use by the existing code is type compatible
and also that an identifier fix up strategy
(as is common in source code transplantation)
is needed.

\vspace{2ex}
\noindent
XML files and test suites are available via 
\url{https://github.com/wblangdon/GPengine_eval_AVX512}

\section*{Acknowledgment}
\noindent
I would like to thank 
Aymeric Blot for help with Magpie,
especially the new version with support for transplanting
via ingredient\_files
and earlier anonymous reviewers.

\bibliographystyle{IEEEtran}

\bibliography{eval_avx512}

\end{document}

%% file: graph512/eval2latex.tex
\begin{tabular}{@{}cc*{16}{r}@{}}
\multicolumn{18}{@{}l}{Program}\\
\hline
1 &
$x$ &
170 &255 &243 &221 &150 &130 &255 &154 &  4 &200 & 96 & 17 &232 &202 & 99 &213 \\
1 &
$y$ &
  0 &  1 &  1 &  5 &  2 &  0 &  1 &  1 & 40 &  0 &122 &  0 &  1 &  0 &  0 &  0 \\
\multicolumn{2}{@{}l}{protected division $x/y$} &
  0 &255 &243 & 44 & 75 &  0 &255 &154 &  0 &  0 &  0 &  0 &232 &  0 &  0 &  0 \\
\hline
1 &
$x$ &
111 &255 & 53 & 20 & 28 &216 & 20 &169 &116 & 63 &160 &248 &217 & 82 &255 &255 \\
1 &
$y$ &
  0 &  1 &  0 &189 &127 &  2 & 79 &  1 &  0 &  0 &  2 &236 &  8 &  0 &  1 &  1 \\
\multicolumn{2}{@{}l}{protected division $x/y$} &
  0 &255 &  0 &  0 &  0 &108 &  0 &169 &  0 &  0 & 80 &  1 & 27 &  0 &255 &255 \\
\hline
1 &
$x$ &
166 &118 &130 &125 &255 &250 &255 &205 &198 & 11 &224 &191 &246 &130 & 91 &240 \\
1 &
$y$ &
  0 &  0 &  1 &112 &  1 &  3 &  1 &  0 &  1 &  0 &  3 & 25 &128 &  3 &  0 &  2 \\
\multicolumn{2}{@{}l}{protected division $x/y$} &
  0 &  0 &130 &  1 &255 & 83 &255 &  0 &198 &  0 & 74 &  7 &  1 & 43 &  0 &120 \\
\hline
1 &
$x$ &
232 & 28 &130 &216 & 12 &231 &227 &196 &115 &186 &151 &161 &219 &204 & 57 &185 \\
1 &
$y$ &
  0 & 54 &  1 &  2 &171 &  2 &  2 &125 &  0 &  1 &  0 &  0 &  0 &  0 & 32 &  2 \\
\multicolumn{2}{@{}l}{protected division $x/y$} &
  0 &  0 &130 &108 &  0 &115 &113 &  1 &  0 &186 &  0 &  0 &  0 &  0 &  1 & 92 \\
\hline
\hline
2 &
$x$ &
247 &124 & 91 &137 & 51 &176 &240 &200 &221 &196 &141 &196 & 97 &118 &132 &247 \\
2 &
$y$ &
  6 &  0 &  1 &  4 &254 &  0 &232 &  1 &  0 &  0 &  2 &  1 &  0 &  1 &247 &236 \\
\multicolumn{2}{@{}l}{protected division $x/y$} &
 41 &  0 & 91 & 34 &  0 &  0 &  1 &200 &  0 &  0 & 70 &196 &  0 &118 &  0 &  1 \\
\hline
2 &
$x$ &
134 & 42 &198 &123 & 96 &184 &244 & 53 &227 & 48 &255 & 29 &178 &202 &255 & 81 \\
2 &
$y$ &
  1 &  0 &  0 & 62 &  0 &  1 &  2 &  0 &  0 & 26 &  1 &  0 &  2 &  0 &  1 &  0 \\
\multicolumn{2}{@{}l}{protected division $x/y$} &
134 &  0 &  0 &  1 &  0 &184 &122 &  0 &  0 &  1 &255 &  0 & 89 &  0 &255 &  0 \\
\hline
2 &
$x$ &
185 & 55 & 48 & 93 & 33 &255 &203 &255 &225 &247 &110 & 63 &103 &  7 &122 &255 \\
2 &
$y$ &
  0 &  0 &  0 &  1 &  0 &  1 &  0 &  1 &190 &  0 &  0 &  3 &128 &  0 &  0 &  1 \\
\multicolumn{2}{@{}l}{protected division $x/y$} &
  0 &  0 &  0 & 93 &  0 &255 &  0 &255 &  1 &  0 &  0 & 21 &  0 &  0 &  0 &255 \\
\hline
2 &
$x$ &
186 & 17 &188 & 48 &222 &120 &163 &155 & 26 &125 & 22 & 32 & 57 &159 & 14 & 92 \\
2 &
$y$ &
  3 &  0 &  0 & 91 &  2 &  1 &  0 &  1 &  0 &  0 &  0 &  0 &  0 &  0 &  0 &  0 \\
\multicolumn{2}{@{}l}{protected division $x/y$} &
 62 &  0 &  0 &  0 &111 &120 &  0 &155 &  0 &  0 &  0 &  0 &  0 &  0 &  0 &  0 \\
\hline
\hline
3 &
$x$ &
105 & 76 & 30 &209 &231 & 72 & 94 & 58 &254 &116 &174 & 80 & 48 &124 & 63 &194 \\
3 &
$y$ &
  1 &  0 &  0 &  0 &  0 &  1 &  1 &  0 &  0 &  0 &  1 &  1 &  0 &  0 & 34 &  0 \\
\multicolumn{2}{@{}l}{protected division $x/y$} &
105 &  0 &  0 &  0 &  0 & 72 & 94 &  0 &  0 &  0 &174 & 80 &  0 &  0 &  1 &  0 \\
\hline
3 &
$x$ &
249 & 18 & 26 &248 &199 &140 &149 & 57 &213 &131 &240 &131 & 28 & 22 &181 &242 \\
3 &
$y$ &
 12 &  0 &  0 &  3 &  0 &  0 &  0 &  0 &  0 &  1 &  0 &  0 &  0 & 35 &  0 &  0 \\
\multicolumn{2}{@{}l}{protected division $x/y$} &
 20 &  0 &  0 & 82 &  0 &  0 &  0 &  0 &  0 &131 &  0 &  0 &  0 &  0 &  0 &  0 \\
\hline
3 &
$x$ &
 42 &241 &163 &183 &156 &175 &136 &233 &108 & 48 &  1 &201 &152 & 23 &  5 &134 \\
3 &
$y$ &
 98 &  2 &  1 &  1 &  1 &  1 &  0 &  0 &  0 &  0 &  0 &  0 & 80 &  0 &  0 &  0 \\
\multicolumn{2}{@{}l}{protected division $x/y$} &
  0 &120 &163 &183 &156 &175 &  0 &  0 &  0 &  0 &  0 &  0 &  1 &  0 &  0 &  0 \\
\hline
3 &
$x$ &
144 & 38 &203 & 66 &179 &255 &194 &107 & 23 &221 &229 &225 &148 &101 & 38 &121 \\
3 &
$y$ &
  1 &  0 &  0 &  0 &  0 &  1 &  0 &125 &  0 & 10 &  2 &  0 &  0 &  0 &  0 &  0 \\
\multicolumn{2}{@{}l}{protected division $x/y$} &
144 &  0 &  0 &  0 &  0 &255 &  0 &  0 &  0 & 22 &114 &  0 &  0 &  0 &  0 &  0 \\
\hline
\hline
4 &
$x$ &
 72 &254 &209 &201 &218 &229 &  7 &187 &218 &223 &255 &255 & 40 &184 & 41 &160 \\
4 &
$y$ &
  0 &  4 &  1 &  0 &  0 &  0 &164 &154 &  0 &  4 &  1 &  1 &  0 &  0 &  0 &  0 \\
\multicolumn{2}{@{}l}{protected division $x/y$} &
  0 & 63 &209 &  0 &  0 &  0 &  0 &  1 &  0 & 55 &255 &255 &  0 &  0 &  0 &  0 \\
\hline
4 &
$x$ &
212 &185 &189 &152 &127 & 15 &221 &152 & 45 & 25 &111 &173 &226 &217 &125 &103 \\
4 &
$y$ &
  1 &  1 &  1 &  2 &  1 &  0 &  1 &  0 &  0 &  0 &  0 &  0 &209 &  1 &  0 &  0 \\
\multicolumn{2}{@{}l}{protected division $x/y$} &
212 &185 &189 & 76 &127 &  0 &221 &  0 &  0 &  0 &  0 &  0 &  1 &217 &  0 &  0 \\
\hline
4 &
$x$ &
 25 &206 & 80 &139 & 57 &249 &208 &206 &127 &250 & 94 &219 &212 &118 & 40 & 89 \\
4 &
$y$ &
  0 &  0 &  0 &  0 &  0 &  2 &  2 &  2 &  2 &  4 &  0 &  2 &  5 &  4 &  0 & 45 \\
\multicolumn{2}{@{}l}{protected division $x/y$} &
  0 &  0 &  0 &  0 &  0 &124 &104 &103 & 63 & 62 &  0 &109 & 42 & 29 &  0 &  1 \\
\hline
4 &
$x$ &
162 &177 &245 &176 &238 &174 &228 & 99 &255 &111 &  3 &154 & 13 &  0 &255 &181 \\
4 &
$y$ &
  0 &168 &  0 &  0 &  0 &162 &  3 &202 &  1 &  2 &  0 &  0 & 13 &  0 &  1 &  2 \\
\multicolumn{2}{@{}l}{protected division $x/y$} &
  0 &  1 &  0 &  0 &  0 &  1 & 76 &  0 &255 & 55 &  0 &  0 &  1 &  0 &255 & 90 \\
\hline
\hline
\end{tabular}

%% file: graph512/eval_progs.tex
\begin{tabular}{@{}ccl@{}}
\multicolumn{2}{@{}l}{Program} & Instruction \\
1 & 0 & R5=R0/R4 \\
  & 1 & R6=R4/126 \\
  & 2 & R6=R0*128 \\
  & 3 & R1=R5/118 \\
\\
2 & 0 & R4=R2/R5 \\
  & 1 & R5=R5-101 \\
  & 2 & R5=R4-R4 \\
  & 3 & R1=R4/R4 \\
\\
3 & 0 & R6=R1/R5 \\
  & 1 & R4=R1+119 \\
  & 2 & R6=R5+60 \\
  & 3 & R7=R5-61 \\
\\
4 & 0 & R3=R3/R6 \\
  & 1 & R6=R3*R3 \\
  & 2 & R1=R6*65 \\
  & 3 & R0=R3-112 \\
\end{tabular}

%% file: graph512/count.tex
\begin{tabular}{@{}lrrr|rl@{}}
Register size & 8 bits & 16 & 32 & 8 bits &
\\\cline{1-5}
Magpie cache        & 15485 &    70 &    76 & 31986 &Section \ref{p.cache}\\
Compilation error   & 12854 &   456 &   522 & 29679 &Section \ref{sec:gcc}\\
Object not changed  &   626 &     3 &     3 &    23 &Section~\ref{sec:equiv_mutants}\\
Run time error      &  4237 &   333 &   282 & 22626 &Secs.\ \ref{sec:fitness} and \ref{sec:runtime_err}\\
All tests past      & 63913 &   580 &   560 & 15687 &Section \ref{sec:better_code}\\
RUN\_TIMEOUT & 0 & 0 & 0 & 0 &
\\
WARMUP & 3 & & & 3 &
\end{tabular}

%% file: graph512/count_xml32.tex
\begin{tabular}{@{}lcrrrrr|r@{}} 
         &     & eval.cpp & regsize & ev diffs & diffs & Guide &  total \\ 
eval.cpp &  ok & 53.3\% &        &  8.1\% &  8.1\% &  4.4\% & 73.9\% \\ 
         & err & 11.8\% &        &  3.0\% &  3.1\% &  3.1\% & 21.0\% \\ 
regsize  &  ok &        &  3.2\% &        &        &        &  3.2\% \\ 
         & err &        &  1.9\% &        &        &        &  1.9\% \\ 
\end{tabular} 

%% file: graph512/count_xml8.tex
\begin{tabular}{lcrrrrr|r} 
         &     & eval.cpp &  & ev diffs & diffs & Guide &  total \\ 
eval.cpp &  ok & 17.4\% &        &  2.5\% &  2.9\% &  0.2\% & 23.1\% \\ 
         & err & 48.6\% &        &  9.9\% &  9.9\% &  8.6\% & 76.9\% \\ 
\end{tabular} 

%% file: graph512/count_gcc_errors32.tex
\begin{tabular}{@{}rrp{2.75in}}
44151 &  63.83\% & 
{\em id} was not declared in this scope \\
12854 &  18.58\% & 
{\em id} was not declared in this scope; did you mean {\em id}? \\
 2920 &   4.22\% & 
redeclaration of {\em id} \\
 2467 &   3.57\% & 
{\em id} without a previous {\em id} \\
 1351 &   1.95\% & 
conflicting declaration {\em id} \\
 1117 &   1.61\% & 
cannot convert {\em id} to {\em id} \\
  913 &   1.32\% & 
expected primary-expression before ? token \\
  627 &   0.91\% & 
cannot convert a value of type {\em id} to vector type {\em id} which has different size \\
  509 &   0.74\% & 
the last argument must be scale 1, 2, 4, 8 \\
  470 &   0.68\% & 
break statement not within loop or switch \\
  387 &   0.56\% & 
cannot convert {\em id} to {\em id} in return \\
  298 &   0.43\% & 
expected primary-expression before {\em id} \\
  193 &   0.28\% & 
unterminated argument list invoking macro "\_mm256\_i32gather\_epi32" \\
   88 &   0.13\% & 
unterminated \#ifndef \\
   86 &   0.12\% & 
expected initializer before {\em id} \\
   79 &   0.11\% & 
unterminated \#else \\
   78 &   0.11\% & 
narrowing conversion of nnn from {\em id} to {\em id} [-Wnarrowing] \\
   78 &   0.11\% & 
a function-definition is not allowed here before {\em id} \\
   73 &   0.11\% & 
decrement of read-only variable {\em id} \\
   65 &   0.09\% & 
jump to case label \\
   52 &   0.08\% & 
\#endif without \#if \\
   44 &   0.06\% & 
invalid types {\em id} for array subscript \\
   32 &   0.05\% & 
unterminated \#if \\
   31 &   0.04\% & 
unable to find numeric literal operator ‘operator""-1’ \\
   31 &   0.04\% & 
unable to find numeric literal operator ‘operator""+1’ \\
   27 &   0.04\% & 
\#else without \#if \\
   24 &   0.03\% & 
no matching function for call to ‘InstrReg32(const OP\&, uint32\_t*\&, int\&)’ \\
   15 &   0.02\% & 
cannot convert ‘const uint16\_t*’ {aka ‘const short unsigned int*’} to ‘retval*’ {aka ‘unsigned char*’} in initialization \\
   11 &   0.02\% & 
declaration of ‘retval* reg’ shadows a parameter \\
   10 &   0.01\% & 
no matching function for call to ‘InstrArg32(const OP\&, uint32\_t*\&, int\&)’ \\
   10 &   0.01\% & 
\#else after \#else \\
    9 &   0.01\% & 
unterminated \#ifdef \\
    8 &   0.01\% & 
cannot convert ‘uint16\_t*’ {aka ‘short unsigned int*’} to ‘const uint8\_t*’ {aka ‘const unsigned char*’} \\
    6 &   0.01\% & 
inlining failed in call to {\em id}: target specific option mismatch \\
    6 &   0.01\% & 
expected ? before ? token \\
    6 &   0.01\% & 
declaration of ‘retval reg [512]’ shadows a parameter \\
    6 &   0.01\% & 
cannot convert ‘uint32\_t*’ {aka ‘unsigned int*’} to ‘const uint8\_t*’ {aka ‘const unsigned char*’} \\
    6 &   0.01\% & 
cannot convert {\em id} to {\em id} {aka {\em id}} \\
    5 &   0.01\% & 
invalid operands of types ‘int [16]’ and {\em id} to binary ‘operator*’ \\
    5 &   0.01\% & 
expected ? before {\em id} \\
    5 &   0.01\% & 
cannot convert ‘const uint32\_t*’ {aka ‘const unsigned int*’} to ‘retval*’ {aka ‘unsigned char*’} in initialization \\
    3 &   0.00\% & 
cannot convert ‘uint16\_t*’ {aka ‘short unsigned int*’} to ‘retval*’ {aka ‘unsigned char*’} in initialization \\
    2 &   0.00\% & 
size ‘12297829382473034240’ of array {\em id} exceeds maximum object size ‘9223372036854775807’ \\
    2 &   0.00\% & 
no matching function for call to ‘InstrReg32(const OP\&, uint32\_t*\&, int)’ \\
    2 &   0.00\% & 
cannot convert ‘uint32\_t*’ {aka ‘unsigned int*’} to ‘retval*’ {aka ‘unsigned char*’} in initialization \\
    2 &   0.00\% & 
cannot convert a vector of type {\em id} to type {\em id} {aka {\em id}} which has different size \\
    2 &   0.00\% & 
cannot convert a vector of type {\em id} to type {\em id} which has different size \\
    1 &   0.00\% & 
size ‘17216961135462247936’ of array {\em id} exceeds maximum object size ‘9223372036854775807’ \\
    1 &   0.00\% & 
redeclaration of ‘retval reg [512]’ \\
    1 &   0.00\% & 
cannot convert {\em id} to {\em id} in initialization \\
    1 &   0.00\% & 
assignment of read-only variable {\em id} \\
\end{tabular}

%% file: graph512/count_gcc_errors8.tex
\begin{tabular}{@{}rrp{2.75in}}
93406 &  62.94\% & 
{\em id} was not declared in this scope \\
25094 &  16.91\% & 
{\em id} was not declared in this scope; did you mean {\em id}? \\
 6947 &   4.68\% & 
redeclaration of {\em id} \\
 4318 &   2.91\% & 
{\em id} without a previous {\em id} \\
 3682 &   2.48\% & 
expected primary-expression before ? token \\
 3089 &   2.08\% & 
conflicting declaration {\em id} \\
 2811 &   1.89\% & 
cannot convert {\em id} to {\em id} \\
 1767 &   1.19\% & 
the last argument must be scale 1, 2, 4, 8 \\
 1573 &   1.06\% & 
cannot convert a value of type {\em id} to vector type {\em id} which has different size \\
 1407 &   0.95\% & 
break statement not within loop or switch \\
  828 &   0.56\% & 
narrowing conversion of nnn from {\em id} to {\em id} [-Wnarrowing] \\
  561 &   0.38\% & 
cannot convert {\em id} to {\em id} in return \\
  559 &   0.38\% & 
decrement of read-only variable {\em id} \\
  410 &   0.28\% & 
unterminated argument list invoking macro "\_mm256\_i32gather\_epi32" \\
  348 &   0.23\% & 
unable to find numeric literal operator ‘operator""+1’ \\
  328 &   0.22\% & 
unable to find numeric literal operator ‘operator""-1’ \\
  274 &   0.18\% & 
unterminated \#ifndef \\
  217 &   0.15\% & 
expected initializer before {\em id} \\
  203 &   0.14\% & 
a function-definition is not allowed here before {\em id} \\
  185 &   0.12\% & 
jump to case label \\
  137 &   0.09\% & 
\#endif without \#if \\
   93 &   0.06\% & 
invalid types {\em id} for array subscript \\
   49 &   0.03\% & 
size ‘12297829382473034240’ of array {\em id} exceeds maximum object size ‘9223372036854775807’ \\
   42 &   0.03\% & 
size ‘17216961135462247936’ of array {\em id} exceeds maximum object size ‘9223372036854775807’ \\
   27 &   0.02\% & 
cannot convert {\em id} to {\em id} {aka {\em id}} \\
   14 &   0.01\% & 
expected ? before {\em id} \\
   11 &   0.01\% & 
inlining failed in call to {\em id}: target specific option mismatch \\
   10 &   0.01\% & 
size of array {\em id} is not an integral constant-expression \\
    5 &   0.00\% & 
invalid operands of types ‘int [16]’ and {\em id} to binary ‘operator*’ \\
    4 &   0.00\% & 
cannot convert {\em id} to {\em id} in initialization \\
    1 &   0.00\% & 
unable to find numeric literal operator ‘operator""+256’ \\
    1 &   0.00\% & 
size ‘17216961135462248174’ of array {\em id} exceeds maximum object size ‘9223372036854775807’ \\
    1 &   0.00\% & 
lvalue required as decrement operand \\
    1 &   0.00\% & 
conflicting declaration ‘uint32\_t tmp [8]’ \\
    1 &   0.00\% & 
conflicting declaration ‘uint32\_t out [8]’ \\
\end{tabular}

%% file: graph512/run_time_errors32.tex
\begin{tabular}{@{}rlrrlp{\temp}@{}}
\multicolumn{2}{c}{Status}&
\multicolumn{2}{c}{fractions}&\\
  0&        & 325270 & 93\%        &             & ok                               \\ 
  1&        &  14740 &  4\%        & 61\%        & Ran ok but gave erroneous outputs\\ 
139& SIGSEGV&   6330 &  2\%        & 26\%        & memory segmentation violation    \\ 
  4&        &   2275 & 0.7\%       &  9\%        & registers corrrpted              \\ 
136&  SIGFPE&    581 & 0.2\%       &  2\%        & divide by zero?                  \\ 
134& SIGABRT&    206 & 0.1\%       & 0.9\%       & registers index error?           \\ 
  3&        &     10 & $3\ 10^{-5}$& $4\ 10^{-4}$& registers padding overwritten    \\ 
  9&   EBADF&      1 & $3\ 10^{-6}$& $4\ 10^{-5}$& perf Bad file descriptor         \\ 
135&  SIGEMT&      1 & $3\ 10^{-6}$& $4\ 10^{-5}$& registers index error            \\ 
\end{tabular}

%% file: graph512/run_time_errors8.tex
\begin{tabular}{@{}rlrrlp{\temp}@{}}
\multicolumn{2}{c}{Status}&
\multicolumn{2}{c}{fractions}&\\
  0&        &  78435 & 41\%        &             & ok                               \\ 
  1&        &  72034 & 38\%        & 64\%        & Ran ok but gave erroneous outputs\\ 
139& SIGSEGV&  29343 & 15\%        & 26\%        & memory segmentation violation    \\ 
  4&        &   6469 &  3\%        &  6\%        & registers corrrpted              \\ 
136&  SIGFPE&   2604 &  1\%        &  2\%        & divide by zero?                  \\ 
134& SIGABRT&   1608 & 0.8\%       &  1\%        & registers index error?           \\ 
  3&        &     28 & $1\ 10^{-4}$& $2\ 10^{-4}$& registers padding overwritten    \\ 
135&  SIGEMT&      1 & $5\ 10^{-6}$& $9\ 10^{-6}$& registers index error            \\ 
\end{tabular}